\documentclass[twoside,journal]{IEEEtran}

\IEEEoverridecommandlockouts                              

\usepackage{cite}
\usepackage{amsmath,amssymb,amsfonts}
\usepackage{algorithmic}
\usepackage{graphicx}
\usepackage{textcomp}
\usepackage{xcolor}

\newcommand{\rev}[1]{{\color{black}#1}}

\usepackage{pifont}
\newcommand{\cmark}{\ding{51}}%
\newcommand{\xmark}{\ding{55}}%

\usepackage{sci}
\usepackage{hhline}
\usepackage{caption} 
\usepackage{array}

\usepackage[breaklinks=true,letterpaper=true,colorlinks,bookmarks=false]{hyperref}

\begin{document}

\title{
Spatiotemporal Relationship Reasoning for Pedestrian Intent Prediction
}

\author{Bingbin Liu$^1$, Ehsan Adeli$^{1}$, Zhangjie Cao$^1$, Kuan-Hui Lee$^2$, \\ Abhijeet Shenoi$^1$, Adrien Gaidon$^2$, Juan Carlos Niebles$^1$ \\
\IEEEauthorblockA{
\textit{$^1$Department of Computer Science, Stanford University, Stanford, CA 94305, USA}\\\textit{$^2$Toyota Research Institute, Los Altos, CA 94022, USA}  \\
\texttt{\footnotesize \{bingbin,\,eadeli,\,caozj,\,ashenoi,\,jniebles\}@cs.stanford.edu} \quad \texttt{\footnotesize \{kuan.lee,\,adrien.gaidon\}@tri.global}}
}



\maketitle

\begin{abstract}
\rev{Reasoning over visual data is a desirable capability for robotics and vision-based applications. Such reasoning enables forecasting the next events or actions in videos. In recent years, various} models have been developed based on convolution operations for prediction or forecasting, but they lack the ability to reason over spatiotemporal data and infer the relationships of different objects in the scene. In this paper, we present a framework based on graph convolution to uncover the spatiotemporal relationships in the scene for reasoning about pedestrian intent. A scene graph is built on top of segmented object instances within and across video frames. Pedestrian intent, defined as the future action of crossing or not-crossing the street, is very crucial piece of information for autonomous vehicles to navigate safely and more smoothly. We approach the problem of intent prediction from two different perspectives and anticipate the intention-to-cross within both pedestrian-centric and location-centric scenarios. In addition, we introduce a new dataset designed specifically for autonomous-driving scenarios in areas with dense pedestrian populations: the Stanford-TRI Intent Prediction (STIP) dataset. Our experiments on STIP and another benchmark dataset show that our graph modeling framework is able to predict the intention-to-cross of the pedestrians with an accuracy of 79.10\% on STIP and 79.28\% on \rev{Joint Attention for Autonomous Driving (JAAD) dataset up to one second earlier than when the actual crossing happens. These results outperform baseline and previous work. Please refer to \url{http://stip.stanford.edu/} for the dataset and code.} 
\end{abstract}


\section{Introduction}
While driving, humans take important and intuitive decisions to achieve safe and smooth navigation. These decisions are ramifications of sequences of actions and interactions with others in the scene. Human drivers can perceive the scene and anticipate if a pedestrian intends to cross the street or not. This is a simple, yet useful piece of information for deciding the next actions to take (\eg, slow down, speed up, or stop). 
It can be made possible through inferring the interdependent interactions among pedestrians and with other items in the scene, like vehicles or traffic lights. Machines, on the other hand, lack the ability to read human judgments with the subtle gestures and interactions they make. This makes autonomous vehicles very conservative and can be nauseating for the riders and revolting for others on the road. 

\begin{figure}[t]
    \centering
    \includegraphics[width=\linewidth]{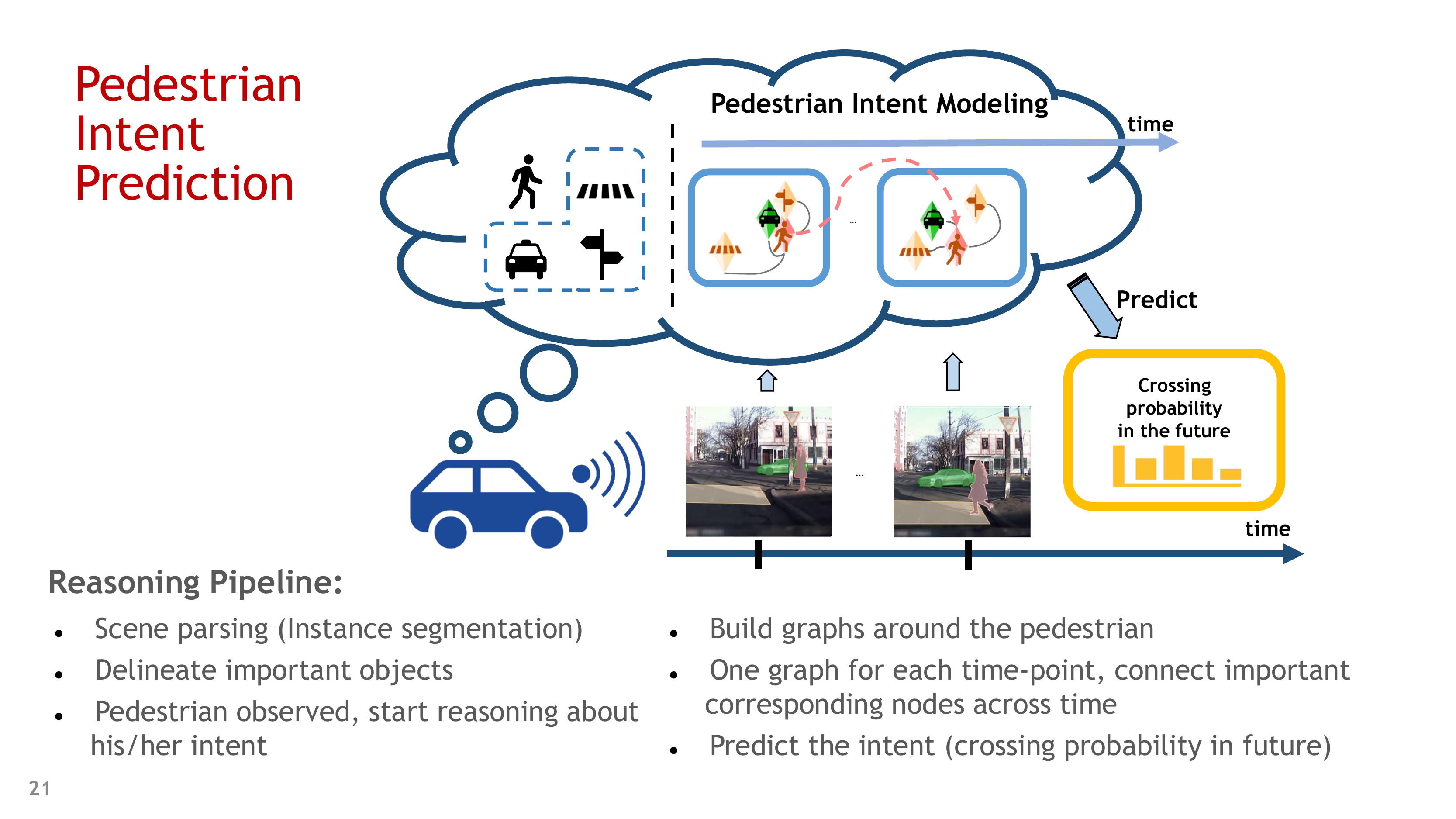}
    \caption{We propose a model for pedestrian intention prediction that could be integrated into a self-driving system. For a car equipped with a front-view camera that continuously captures imagery of the environment, our model parses the visual input into a \textit{pedestrian-centric graph} to understand the relationships among observed entities. These relationships capture the spatiotemporal context of the surroundings, which is essential in predicting future behaviors of the pedestrian, as evidenced by our experimental results.}
    \label{fig:pull}
\end{figure}


Developing algorithms that read pedestrian instincts and make judgments based on them requires reasoning about the objects in the scene and how they interact, \ie, visual relationships. 
Most previous works modeling pedestrians focus on pedestrian detection and tracking \cite{cite:Springer14VB,cite:TPAMI18Towards,cite:ICIP17Simple,cite:TPAMI14VirtualReal,cite:CVPR18OcclusionSingle,cite:ECCV18SingleAsymLF,cite:ECCV18Crowd,cite:CVPR18Crowd} or behavior analysis \cite{cite:ITSC14BLAC,cite:ECCV14CBP}. Although they have obtained convincing performance on several benchmarks, completion of such tasks is not enough for human-like driving. 
On the other hand, trajectory prediction  \cite{cite:NIPS17CASNSC,cite:ArxivLTRPM,cite:ICRA16IALT,cite:Arxiv18Transferable,mangalam2020disentangling,cite:CVPR18CrowdTraj,alahi2016social,sadeghian2018sophie} addresses the problem to some extent by predicting the potential future position of the pedestrian. But predicting trajectories with high confidence long-enough into the future is a very challenging task as many different and subtle factors change the trajectories of pedestrians. 
Contrarily, pedestrian intents are high-level semantic cues that can dramatically influence autonomous driving systems. Intent can be defined as their future and immediate action of whether they will cross the street or not. Anticipation and early prediction of pedestrian intent will help to demonstrate safer and more smooth driving behaviors. 

\rev{Recent work \cite{cite:ER13IAPA,cite:ITSC14MCBM,cite:IVS182DPose,seo2018structured,rasouli2019pie} introduced pedestrian intent prediction and have typically tackled the problem by observing} pedestrian-specific features such as location, velocity, and pose. Although these cues can contribute to inferring pedestrian intent, they ignore context and pedestrian interactions with elements in the scene, such as other pedestrians, vehicles, traffic signs, lights, and environmental factors, \eg, zebra-crossings. We argue that such interactions can be uncovered through reasoning over objects  relationships through time. Therefore, we explore graph-based spatiotemporal modeling. 

In this paper, we propose an approach for the pedestrian intent prediction problem based on visual relationship reasoning (Fig. \ref{fig:pull}). To this end, we build a pedestrian-centric dynamic scene graph. We first generate instance segmentation of each frame in the video using a pre-trained off-the-shelf instance segmentation model \cite{li2018learning}. We then extract features from each instance in the image and reason about the relationship between pairs of instances through graph convolution techniques. One graph is defined for each pedestrian instance testifying to his/her intent. The pedestrian node is connected to all other instance nodes as well as a context node, which aggregates all the contextual visual information. To model pedestrian actions and interactions with others through time, we connect pedestrian and context nodes between consecutive frames to further reasoning about the temporal relations. This spatiotemporal modeling allows for capturing the intrinsic scene dynamics encoding the sequence of human subtle judgments and actions, which are very important for inferring the intent. In addition, we study the problem from a different point of view, by building a location-centric graph. In this setting, we predict how likely it is that a pedestrian will show up in front of the autonomous vehicle in the near future. This is critically important knowledge for an autonomous driving agent and our visual relationship reasoning is capable of modeling it. With such spatiotemporal relationship reasoning, we obtain reasonable results for intent prediction, which outperforms all baseline and previous methods. 
Comprehensive description of the related work  are provided in the supplement. 

In summary, the contributions of this work are two-fold: (1) We model the problem of intent prediction via instance-level spatiotemporal relationship reasoning and adopt graph convolution techniques to uncover individual intent; (2) Our modeling involves observing the problem from two different perspectives of pedestrian-centric and location-centric settings, both of which are crucial for autonomous driving applications. In addition, We also introduce a new dataset specifically designed for intent prediction in vehicle-centric view scenes.

\section{Related Work}
\noindent\textbf{Pedestrian Detection and Tracking} are basic steps for reasoning about the pedestrian intent. Previous work about vision-based pedestrian protection systems \cite{cite:Springer14VB} provides a thorough investigation of such methods based on shallow learning. Recently, various deep learning methods are proposed for single-stage detection \cite{cite:CVPR18OcclusionSingle,cite:ECCV18SingleAsymLF}, detection in a crowd \cite{cite:ECCV18Crowd,cite:CVPR18Crowd}, and detection at the presence of occlusion \cite{cite:CVPR18OcclusionSingle,cite:ECCV18OCcclusion,cite:CVPR18Occlusion}; all these methods obtain prominent accuracies for pedestrian detection. For pedestrian tracking, multi-person tracking methods \cite{cite:ECCV16MultiTrack,cite:CVPR17MultiTrack} are proposed to track every person in a crowded scene. Recently, tracking problems are simultaneously solved with pose estimation \cite{cite:CVPR17MultiTrackPose,cite:ECCV18MultiTrackPose,cite:Arxiv18TrackPose} and person re-identification \cite{cite:CVPR17MultiTrackReid,cite:CVPR18MultiTrackReid} in a multi-task learning paradigm. Given the obtained promising results, we take them for granted and investigate visual reasoning schemes to understand the intrinsic intent of the pedestrians. 

\noindent\textbf{Trajectory Prediction} is another closely-related task for understanding the pedestrian intent. Recent works leverage human dynamics in different forms to predict trajectories. For instance, \cite{cite:ITSC14BLAC} proposes Gaussian Process Dynamical Models based on the action of pedestrians and \cite{cite:ICRA16IALT} uses an intent function with speed, location, and heading direction as input to predict future directions. Other works incorporate environment factors into trajectory prediction \cite{cite:ECCV14CBP,cite:NIPS17CASNSC,cite:ICRA16ADL,cite:Arxiv18Transferable}.
Some other works observe the past trajectories and predict the future. For instance, \cite{cite:ArxivLTRPM} combines inverse reinforcement learning and bi-directional RNN to predict future trajectories. Recently, \cite{cite:CVPR18CrowdTraj} proposed a crowd interaction deep neural network to model the affinity between pedestrians in the feature space mapped by a location encoder and a motion encoder. 
A large body of trajectory prediction methods depends on top-down (bird's eye) view. Among these works, Social LSTM \cite{alahi2016social} incorporates common sense rules and social interactions to predict the trajectories of all pedestrians. Social GAN \cite{gupta2018social} defines a spatial pooling for motion prediction. SoPhie \cite{sadeghian2018sophie} introduces an attentive GAN to predict individual trajectories leveraging the physical Constraints. Although obtained impressive results, these top-down methods pose limitations that make them inapplicable to egocentric applications of self-driving scenarios.

One can argue that if we can accurately predict the pedestrians future trajectories, we already know their intent. This is valid, but trajectory prediction is more complex and requires more annotations and supervision. In addition, it is not a well-defined problem as future trajectories are often very contingent and cannot be predicted long enough into the future with enough certainty. In contrast, we look at the intent of the pedestrians defined in terms of future actions (cross or not cross) based on reasoning over the relationship of the pedestrian(s) and other objects in the scene. 

\noindent\textbf{Pedestrian Intent Prediction} is explored by only a few previous works. For instance, \cite{cite:ER13IAPA} uses LIDAR and camera data to predict pedestrian intent based on location and velocity. Bonnin \etal~\cite{cite:ITSC14MCBM} use context information to calculate predefined crafted features for intent prediction. \cite{lan2014hierarchical} proposes hierarchical movements to represent human action and predict human action from human appearance. \cite{cite:IVS182DPose} extracts features from pedestrian key-points, and integrates features of neighboring frames to predict whether the pedestrian will cross. \rev{In other works, \cite{seo2018structured} introduces a sequence model and \cite{rasouli2019pie}, a concurrent work with us, a dataset for this task.} These works only use features from human without the context information in the scene, while our model leverages temporal connected spatial graph to incorporate relations between objects in the scene to encode the dynamics context information. This facilitates realistic visual reasoning to infer the intent, even in complex scenes. Recent work \cite{xie2017learning,wei2018and} consider context information but they require additional moralities or constraints, which is not common across datasets, such as depth modality for \cite{wei2018and} and bird's-eye view for \cite{xie2017learning}. Whereas we only use raw video frames as the input.

\noindent\textbf{Action Anticipation and Early Prediction} methods can be considered as the most relevant methodological ramifications of intent understanding. Among these works, \cite{abu2018will,shi2018action} learns models to anticipate the next action by looking at the sequence of previous actions. Other works build spatiotemporal graphs \cite{rhinehart2017first} for first-person action forecasting, or use object affordances \cite{koppula2016anticipating} and reinforcement learning \cite{chen2018part} for early action prediction. In contrast, instead of only looking at the data to build a data-driven forecasting model, we build an agent-centric model that can reason on the scene and estimate the likelihoods of crossing or not-crossing. 

\noindent\textbf{Scene Graph Parsing and Visual Reasoning} 
Modeling spatial and temporal context with graph have been width explored recently. There are works focusing on toy datasets \cite{battaglia2016interaction,van2018relational}. In the case of real scene, scene graphs have been a topic of interest for understating the relationships between objects encoding rich semantic information about the scene \cite{xu2017scene}. The previous work generated scene graphs using global context \cite{zellers2018neural}, relationship proposal networks \cite{yang2018graph}, conditional random fields \cite{cong2018scene}, iterative message passing \cite{xu2017scene} or recurrent neural network \cite{ibrahim2016hierarchical}. Such graphs built on top of visual scenes were used for various applications, including image generation \cite{johnson2018image}, action recognition \cite{wang2018videos}, trajectory prediction \cite{kipf2018neural} and visual question answering \cite{teney2017graph}. However, one of their main usages is reasoning about the scene, as they outline a structured representation of the image content. Among these works, \cite{shi2018explainable} uses scene graphs for explainable and explicit reasoning with structured knowledge. Aditya \etal~\cite{aditya2018image} use directed and labeled scene description graph for reasoning in image captioning, retrieval, and visual question answering applications. In another recent work, \cite{chen2019graph} introduces a method for globally reasoning over regional relations in a single image. In contrast to the previous work, we build agent-centric (\eg, pedestrian-centric) graphs to depict the scene from the agent's point of view. We use a context node to cope with varying number of objects, which relaxes the constant graph size constraints required by several previous works \cite{battaglia2016interaction,van2018relational,kipf2018neural,jain2016structural,qi2018learning,shu2017cern}.  Furthermore, instead of creating one single scene graph, we build a graph for each time-point and connect the important nodes across different times to encode the temporal dynamics (denoted by temporal connections). We show that these two characteristics can reveal pedestrian intent through reasoning on the spatiotemporal sequence of visual data. 


\section{Method}\label{sec:method}
\begin{figure*}[ht]
    \centering
    \includegraphics[width=\linewidth]{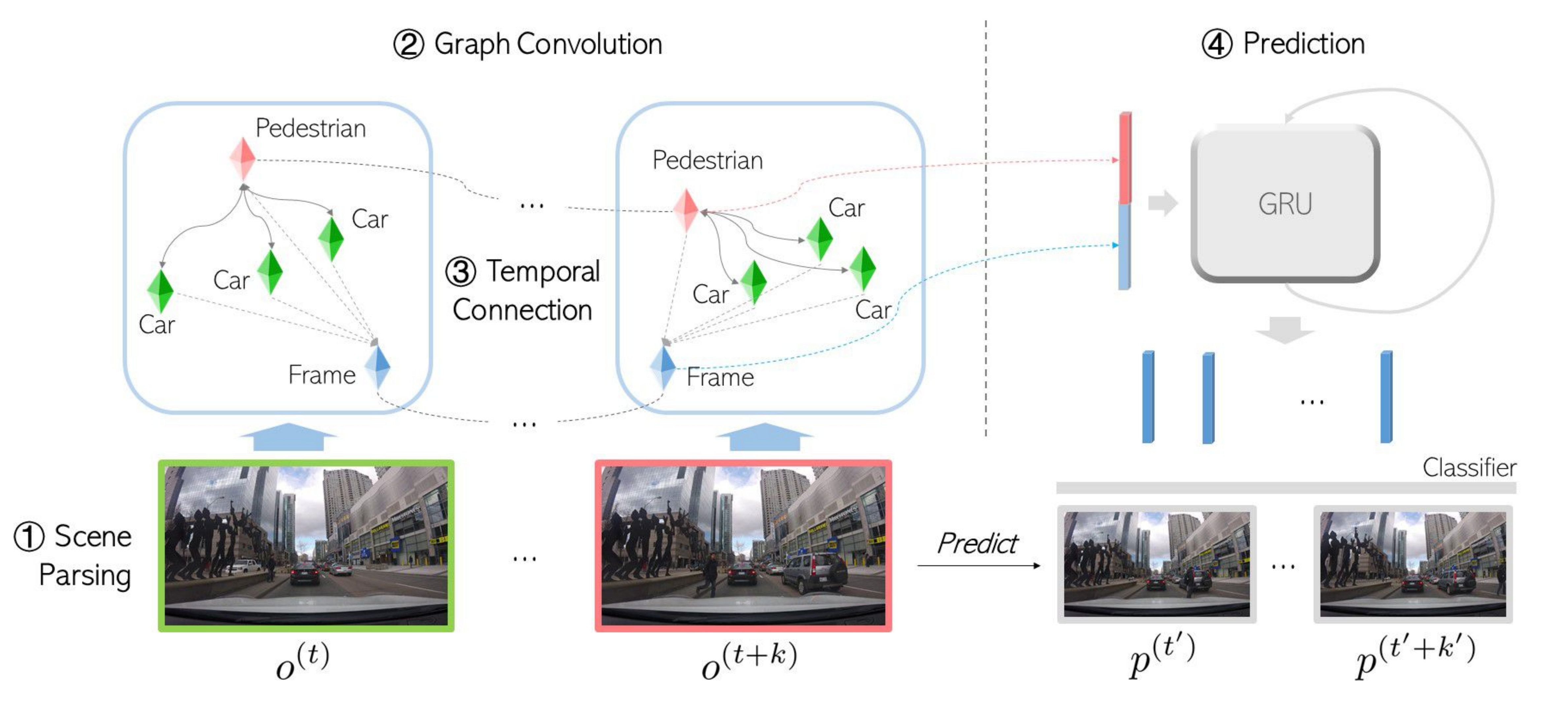}
    \caption{Overview of the model. With a sequence of input frames ($\{o^{(t)}\}$), the model observes whether a pedestrian is crossing (frame in red) or not (frame in green) by capturing the relationship between the pedestrian and the surrounding. It then predicts the future crossing likelihood up to a certain temporal horizon ($\{p^{(t')}\}$). Specially, this is achieved with four components: 1) \textbf{Scene Parsing} (Sec. \ref{sec:method_scene}): the input video frames are first parsed into pedestrians (bounding boxes) and objects (binary masks); 2) \textbf{Graph Convolution} (Sec. \ref{sec:method_gcn}): a pedestrian-centric graph is built on each frame to connect a pedestrian with the surrounding; 3) \textbf{Temporal Connection} (Sec. \ref{sec:method_temp}): the pedestrian nodes at each frame and frame-level representations are connected temporally; 4) \textbf{Prediction} (Sec. \ref{sec:method_pred}): using the rich representation on observed frames, the model tries to predict the future crossing behavior of the pedestrian of interest. More details are described in Section \ref{sec:method}.}
    \label{fig:gcn}
\end{figure*}

We propose a model that leverages the spatiotemporal context of the scene to make the prediction. Given a sequence of video frames observed in the past, the model first parses each frame into pedestrian and objects of interests, each of which is encoded to a feature vector (Section \ref{subsec:scene_parsing}). Then for each pedestrian, we construct a pedestrian-centric spatiotemporal graph using these features as node representation and produce a feature vector that encodes both scene context and the temporal history in the observed frame (Section \ref{subsec:graph_conv}). Finally, an RNN is used to predict the behavior of the pedestrian (Section  \ref{subsec:temporal_connection}). Fig. \ref{fig:gcn} shows an overview of the model.

\subsection{Scene Parsing}\label{sec:method_scene}
\label{subsec:scene_parsing}
Self-driving systems equipped with egocentric cameras often need to cope with noisy information coming from busy views. We hence remove irrelevant information by focusing only on the pedestrians, the vehicles, and certain related items in the environment. Table \ref{tab:objects} provides a complete list of objects of interests. We crop out pedestrians using ground truth bounding boxes. The reason of using ground truth is that pedestrian detection can be considered a solved problem given recent progress on person detectors \cite{ren2015faster,lin2017focal,redmon2018yolov3}, and that we would like to better evaluate our proposed method of leveraging spatiotemporal context by isolating the graph component which we will introduce later. Similarly, we use off-the-shelf instance segmentation framework \cite{li2018learning} trained on \cite{neuhold2017mapillary} to obtain the binary object masks. Then for each object, we crop out the union bounding box enclosing both the object and the pedestrian, such that the relative position of the object and the pedestrian can be preserved. Note that the appearance information is thrown away, since it is the location and movement of the object relative to the pedestrian that affects the pedestrian's crossing behavior the most, rather than the exact appearance such as color and texture. This further simplifies the information that the model needs to handle, and was proven to be beneficial in the experiments. Finally, the cropped out pedestrian and the union binary masks are encoded using two separately tuned ResNet-18 backbone.

\begin{table}[]
    \caption{List of objects of interest.}
    \setlength{\tabcolsep}{3pt}
    \label{tab:objects}
    \centering
    \begin{tabular}{l|p{6.5cm}}
        \textbf{Category} & \textbf{Objects} \\ \hline
        Vehicle & bike, bus, car, caravan, motorcycle, trailer, truck, other \\ 
        Road user & bicyclist, motorcyclist, other riders. \\ 
        Environments & crosswalk (plain), crosswalk (zebra), traffic light. \\ 
    \end{tabular}
\end{table}

\subsection{Graph Convolution for Modeling Spatiotemporal Context}\label{sec:method_gcn}
\label{subsec:graph_conv}
The main contribution of this work is to augment the prediction model with context information, including both spatial context from objects in the scene, as well as the temporal context from the history. We hence propose a pedestrian-centric spatiotemporal graph spanning both space and time.

\subsubsection{Pedestrian-Centric Graph}
We use a graph structure to make use of the context information. Intuitively, each pedestrian or object corresponds to a graph node and the edges reflects the relationship strength between the two connected nodes. We define the graph convolution operation \cite{kipf2016semi} as:
\begin{equation}
	Z = AXW,
\end{equation}
where $X$ is a matrix whose rows are feature vectors for the graph nodes, $W$ is the trainable weight of a graph convolution layer, and $A$ is the adjacency matrix of the graph.

Since our goal is to predict the crossing behavior for each pedestrian, we model each pedestrian with a star graph centered at the pedestrian.
We define the edges using information from both the spatial relationship between the pedestrian and objects, as well as the appearance of the pedestrian. \rev{The spatial relationship is a good indicator of the importance of an object; for example, distance of the objects with respect to the pedestrian can play an important role. The appearance of the pedestrian is also considered since it can serve as a strong cue to the pedestrian's intent, which can often be inferred from the head orientation or gazing direction. For example, an object to which the pedestrian is giving high attention should be associated with a heavier edge weight.}

Inspired by \cite{zeng2017agent}, we describe the spatial relationship with a length-8 vector $s$, whose entries include the height and width of the union bounding box, and the differences between box corners and center: $\delta x^{(min)}$, $\delta y^{(min)}$ for the upper left corner, $\delta x^{(max)}$, $\delta y^{(max)}$ for the lower right corner, and $\delta x^{c}$, $\delta y^{c}$ for the center.
We then combine the spatial vector with the feature representation of the pedestrian and use them to calculate the edge weight. The edge weight for the $i_{th}$ object $o_i$ is
\begin{eqnarray}
\begin{aligned}
	s_i &= [\delta x^{(min)}_i, \delta y^{(min)}_i, \delta x^{(max)}_i, \delta y^{(max)}_i, \delta x^c_i, \delta y^c_i, w_i, h_i], \\
	v_i &= [v_a, s_i], \\
	w_i &= \text{sigmoid}(\text{ReLU}(v_i) \cdot \text{ReLU}(v_o)),
\end{aligned}
\end{eqnarray}
where $v_a$ is the feature vector for the pedestrian capturing the appearance, and $v_o$ is the feature vector for the binary mask for object $o_i$.
For a graph with $N$ object nodes, $A$ is a symmetric matrix whose entries are, assuming $i\leq j$,
\begin{equation}
	A[i, j] = \begin{cases}
	1, & i = j, \\
	w_j, & i = 1, j \neq 1, \\
	0, & \text{otherwise}.
	\end{cases}
\end{equation}
where the $1_{st}$ node is the pedestrian and the other $N$ nodes correspond to the objects.

\subsubsection{Location-Centric Prediction}
\begin{figure}
    \centering
    \includegraphics[width=\linewidth]{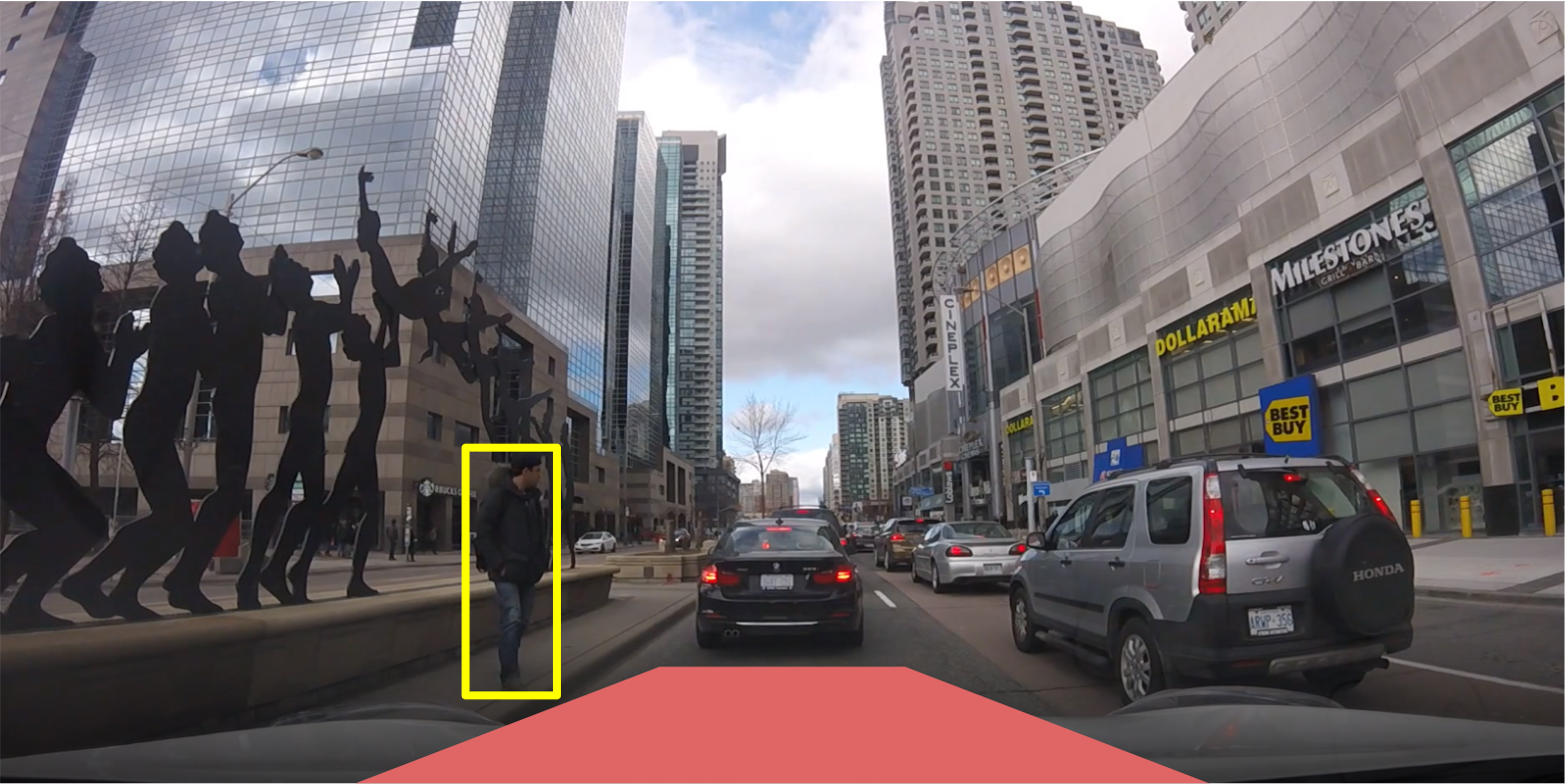}
    \caption{Intent prediction under the location-centric setting, where crossing in the future is equivalently mapped to crossing in the distance at the moment. By focusing on the area that the egocar is going to cover in the near future (highlighted in red), we construct a star graph centering on the egocentric view, with objects and pedestrians in the scene being the nodes.}
    \label{fig:loc_centric}
\end{figure}

We also consider the location-centric setting, where instead of predicting crossing behavior for each pedestrian, we predict the probability that there is someone crossing a designated area. Fig. \ref{fig:loc_centric} shows an illustration of the setting. At each moment, the model focuses on a trapezoid area which can be interpreted as the area that the current car will cover in the near future for which we are predicting. In this way, the problem of predicting whether some pedestrian of interests will cross the road in the next $T$ seconds can equivalently be translated to whether any pedestrians will enter the focused red area that the car will cover in $T$ seconds.

This setting is beneficial especially for busy scenes with numerous pedestrians, since instead of building a graph for each pedestrian, the model focuses only on those that may affect the driving behavior. It not only simplifies the computation but is also more relevant from a control point of view. 

We also modify the pedestrian-centric graph to be location-centric, where the center node encodes the point of view of the ego-car, and the peripheral nodes are other vehicles in the surrounding, riders, traffic signs, as well as the pedestrians. Rather than computing pairwise relative spatial location as part of the edge weight, we embed the egocentric scene and the context objects (including human) into a common embedding space, on which we define the edge weight to be the inner product of the egocentric scene and an object constrained with a sigmoid function.

\subsection{Temporal Connection}\label{sec:method_temp}
\label{subsec:temporal_connection}
The previous pedestrian-centric star graph is constructed on each frame. However, we would like the communication within the graph to consider also the temporal history. In order to encode temporal relation into the node features, we first connect the pedestrian nodes in each frame with a GRU. Modeling the temporal relation in the contextual objects would be slightly trickier since the number of objects present in each frame may vary. However, we do not explicitly draw temporal associations among objects in different frames, since when performed properly, graph convolution would guarantee that information is sufficiently communicated among the nodes, hence the temporal information can be captured by the pedestrian node and pass to the context objects. We will also experimentally show this in the ablation study.

\subsection{Prediction GRU}\label{sec:method_pred}
To leverage the spatiotemporal context, the model performs two layers of graph convolution on each observed frame, where the features for the pedestrian node and the context node are hidden states of the corresponding GRUs. The refined pedestrian and context feature vectors after graph convolution are then concatenated at each frame, and are aggregated by an additional GRU. The last hidden state of this additional GRU is then used for anticipating crossing behavior in the future, which is achieved by a designated prediction GRU.

\section{Experiments}
In this section, we evaluate our method on \rev{two datasets and compare the results with a wide range of baselines. We examine different settings of model structure (ablation studies) as well as choices of the features. We also explore how long in the future can be predicted by expanding the temporal horizon.}  
\subsection{Datasets}
\subsubsection{Joint Attention Autonomous Driving (JAAD) Dataset} JAAD \cite{kotseruba2016joint} dataset focuses on joint attention in the context of autonomous driving in everyday urban setting. It is designed to explore pedestrian and driver behaviors with labels of when the pedestrians cross. 
JAAD contains videos captured with a front-view camera under various scenes, weathers, and lighting conditions. There are 346 videos with 82032 frames, where the length of videos ranges from 60 frames to 930 frames. 
Each video may contain multiple pedestrians with ground truth labels on 2D locations in the frame and 9 actions. We report metrics only on the action of crossing, but all 9 action labels are used for model training. The dataset is split into 250 videos for training and the rest for testing, which correspond to 460 pedestrians and 253 pedestrians respectively. The crossing action is roughly balanced, with 44.30\% frames labeled as crossing and 55.70\% as non-crossing throughout the dataset.

\subsubsection{Stanford-TRI Intent Prediction (STIP) Dataset}
\begin{figure}
    \centering
    \includegraphics[width=\linewidth]{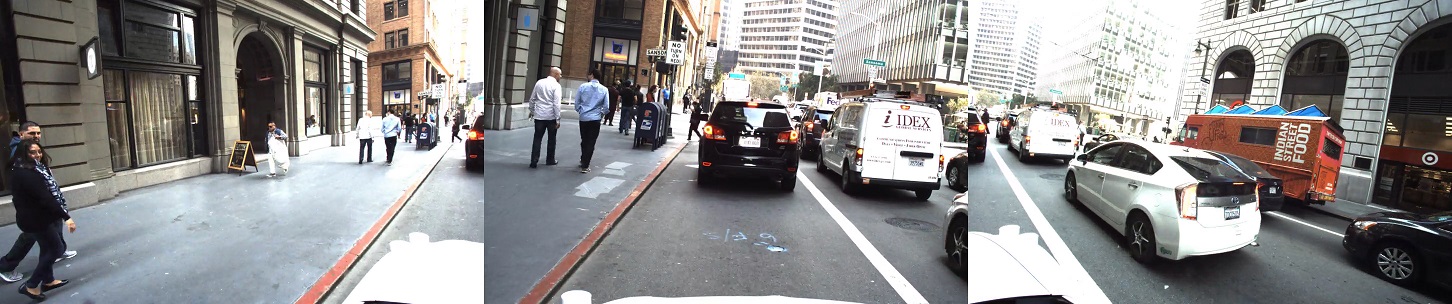}
    \includegraphics[width=\linewidth]{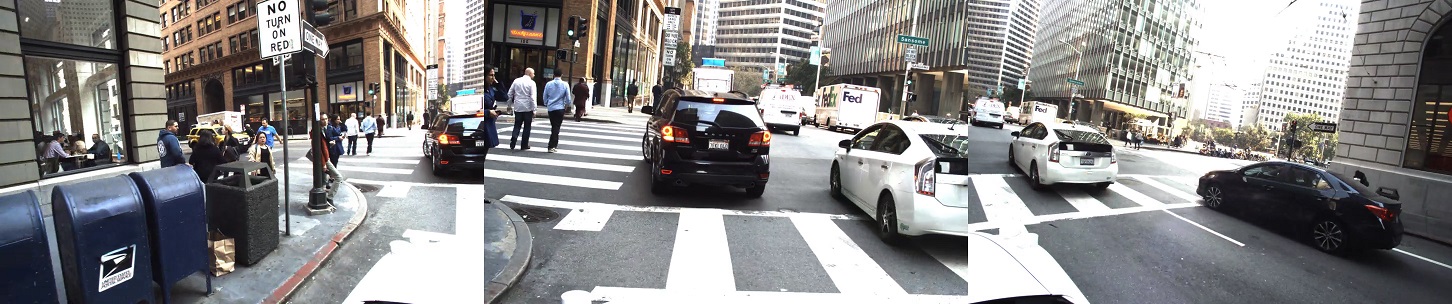}
    \includegraphics[width=\linewidth]{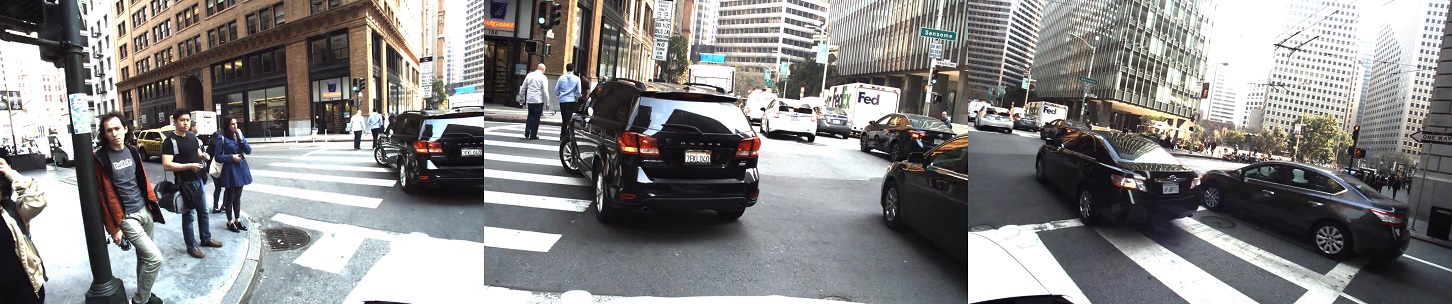}
    \includegraphics[width=\linewidth]{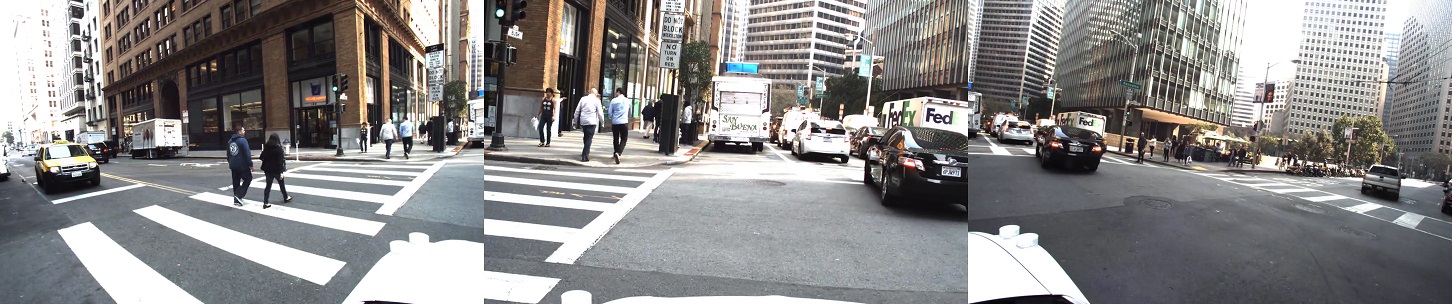}
    \caption{One sample scene with left, front, and right camera views from the STIP dataset. All pedestrains in all views are annotated with bounding boxes and cross/not-cross labels (omitted from this figure for clarity).}
    \label{fig:stipsample}
\end{figure}

In this paper, we introduce a new dataset of driving scenes recorded in dense urban areas in the United States in California and Michigan, at 8 cities under various weather conditions. 
This dataset is created within the context of a collaboration between Stanford University and the Toyota Research Institute (TRI). The dataset contains 923.48 minutes (at 20 fps; 1,108,176 frames in total) of driving scenes with high quality ($1216\times 1936$) recordings. A total of over 350,000 pedestrian boxes were annotated manually at 2 fps. The total number of pedestrian tracks are over 25,000 with a median length of 4 seconds. Each annotated sequence contains video recordings of three cameras simultaneously (left, front, and right). One sample sequence is shown in Fig. \ref{fig:stipsample} \rev{, and Table \ref{tbl:stip} compares STIP with other available autonomous driving datasets. Our dataset has the longest length taken in dense urban areas with largest number of annotated (and interpolated) frames.}

\begin{table}[t]
    \caption{\rev{Comparison of STIP with other datasets. The columns indicate publication year of the dataset, total length of the videos in minutes, number of annotated frames, number of annotated (\& interpolated) pedestrian instances, number of cameras, and finally if the dataset has cross/not cross (C/NC) annotations on each pedestrian at each frame. STIP has 350K annotated (3.5M interpolated) pedestrian instances.}}
    \centering
    \setlength{\tabcolsep}{3pt}
   \rev{ \begin{tabular}{l|cccccc}
         \textbf{Dataset} & \textbf{Year} & \textbf{Len (min)} & \textbf{\#Frames} & \textbf{\#Peds}  & \textbf{\#Cams} & \textbf{C/NC} \\ 
         \hline
         KITTI \cite{geiger2013vision} & 2012 & 90 & 80,000 & 12,000 & 1 & \xmark \\ 
         JAAD \cite{kotseruba2016joint} & 2017 & 46 & 82,000 & 337,000 & 1 & \cmark \\ 
         BDD 100k \cite{yu2018bdd100k} & 2017 & 60,000 & 100,000 & 86,047 & 1 & \xmark  \\
         PedX \cite{kim2019pedx} & 2018 & - & 10,152 & 14,091 & 1 & \xmark \\
         NuScenes \cite{caesar2019nuscenes} & 2019 & 330 & 1,400,000 & 1,400,000 & 6 & \xmark \\
         Ours (STIP) & 2020 & 923.48 & 1,108,176 & 
         3,500,000 & 3 & \cmark \\ 
         \hline
    \end{tabular}}
    \label{tbl:stip}
\end{table}

Given these manually annotated frames, we used instance segmentation results and a JPDA \cite{hamid2015joint} based tracker to interpolate. Post-processing was applied to constrain the tracks to the known, manually annotated, ground truth frames. The resulting trajectories are at 20 fps are prepared for release; however, in this work we report results on the manually annotated 2 fps data. For pedestrian intent prediction, as a preliminary study, we focus on the parts of videos where the car is around an intersection. We manually select front view cameras of 556 video segments which happen in busy intersections. These video segments are then translated to 2525 pedestrians in 102.37 minutes of videos for training, and 823 pedestrians in 23.43 minutes of videos for testing. 

{}


\subsection{\rev{Baseline Models}}
The model takes in $T$ frames as past observation, and predict the per-frame probability of crossing for up to $K$ frames in the future. We report separately the performance averaged over the $K$ future frames, as well as the performance on the $K^\text{th}$ frame, which helps to estimate how the prediction performance scales as the temporal horizon increases.
The following baseline models are used to compare with our proposed model:
    
    \noindent \textit{Pose-based method}: we compared with \cite{cite:IVS182DPose} which predicts the crossing behavior using pose key points. For each frame, the input is a vector containing 18 2D coordinates from the same 9 joints as in their paper, which are the neck, the left/right shoulder, the left/right hip, the left/right knee, and the left/right ankle. The joints are obtained from OpenPose \cite{cao2018openpose} pretrained on COCO \cite{lin2014microsoft} without fine-tuning. We re-do the experiments on JAAD since we are using the newest version of JAAD, different from that used in \cite{cite:IVS182DPose} , and also since we use different numbers of observed and predicted frames.
    
    \noindent \rev{\textit{Pedestrian Locomotion Forecasting (PLF)} \cite{mangalam2020disentangling}: One of the strong baselines for intent prediction is to first predict the trajectories of the pedestrians and then use the predicted trajectory to classify the intent. We use two strong trajectory prediction methods of `Constant Velocity' and PLF. We also use PLF to predict both the future pose and trajectory, based on which we then predict the intent. Our experimental settings are the same as those reported in \cite{mangalam2020disentangling}.}
    
    \noindent \textit{1D CNN} \cite{abu2018will}: predicting the future crossing behavior can be considered as a type of action anticipation. We therefore also include the comparison with the 1D CNN module 
    which predicts the future action labels directly from the observed action labels. This baseline model shares the same backbone as our detection model.
    
    \noindent \textit{Temporal Segment Network (TSN)} \cite{TSN2016ECCV}: we take the model with BN-Inception backbone. 
    pretrained on ImageNet. Since our experiments show that the pedestrian region is better than taking in the entire frame, we use the same pedestrian regions as input for the TSN.
    
    \noindent \textit{Temporal Relation Network (TRN)} \cite{TRN}: we also take the model with BN-Inception backbone pretrained on ImageNet, and takes as input the pedestrian regions. We conduct experiments on both the single-scale and the multi-scale version, where the largest scale $N$ is set as $8$ in the multi-scale version as in the original paper. However, since TRN based its prediction on features from frame tuples, where no frame-wise prediction is available. We only report results on the last prediction setting, where only one final prediction is needed.
    

\begin{table}[t]
    \caption{\textbf{\rev{Accuracy comparison with baseline models}}: \rev{our model (row \#8) outperforms state-of-the-art models based on trajectory prediction (rows \#1\&2), past pose (row \#3), past and predicted future pose (row \#4), action anticipation (row \#5), and early action recognition (rows \#6\&7).}} 
    \label{tab:main}
    \centering
    \begin{tabular}{c|l|c|c}
        \hline
        \# & \textbf{Model} & \textbf{Avg on 1-30 frames} & \textbf{30$^\text{th}$ frame} \\
        \hline
        1 & \rev{Constant velocity} & {70.08\%} & {68.30\%} \\
        2 & \rev{PLF (traj.) \cite{mangalam2020disentangling}} & \rev{69.82\%} & \rev{64.56\%} \\
        \hline
        3 & Pose-based \cite{cite:IVS182DPose} & 67.00\% & 67.00\% \\ 
        4 & \rev{PLF (pose) \cite{mangalam2020disentangling}} & \rev{71.36\%} & \rev{68.25\%} \\
        \hline
        5 & 1D CNN \cite{abu2018will} & 72.78\%  & 69.65\% \\ 
        6 & TSN \cite{TSN2016ECCV} & 67.18\% & 63.64\% \\
        7 & TRN \cite{TRN} & - & 63.74\% \\
        \hline
        8 & Ours & \textbf{79.28\%} & \textbf{76.98\%} \\
        \hline
    \end{tabular}
\end{table}
\subsection{\rev{Results on JAAD Dataset}}
Table \ref{tab:main} shows the results comparing our proposed model with the baselines, where our model outperforms the baselines by a large margin. One possible explanation is that most of the baseline models are geared towards action recognition, early action recognition, or action anticipation. Hence, they are not optimized for the task of predicting pedestrian intention.

The performance of the pose-based model was not as good as we would expect from \cite{cite:IVS182DPose}, which we think may be caused by two reasons: first, the poses output by the pose estimator were not of decent quality due to a more challenging dataset. Compared to \cite{cite:IVS182DPose} which used JAAD v1, the current version of JAAD is of much larger size and more complex. Furthermore, \cite{cite:IVS182DPose} simplifies the task by leaving out the prediction for pedestrians with less than 60 pixels in width.

 
\subsubsection{\rev{Ablation Study}}
We present the results on ablation experiments to analyze the contribution of each model component in isolation, as well as to select the most suitable design choice in terms of both the model structure and use of features.

\noindent\textbf{Choice of Model Structure:}

\textit{Graph-based \vs~Concatenation}: incorporating context information could take up a simple form by directly concatenating the feature vectors of the contextual objects and the pedestrian. Since the number of objects may vary across frames, we represent the context by encoding the union of the binary masks with the ResNet backbone. Note that we chose to encode the binary masks rather than the original frame in RGB, since the latter was shown to give poorer performances. These preliminary experiments help us decide on parsing the scene with a segmentor, which reduces the noises in the scene while preserving the semantic information and spatial relation. We use two GRUs to model temporal connectivity, one on pedestrian features and one on the concatenated frame-level features, which is the same setting as our full model so as to have a fair comparison. The results of this concatenation model are shown in the first row of Table \ref{tab:ablation_struct}.

\textit{Variation on graph structures}: the variations are defined in terms of the number of convolution operations performed, and the connectivity of the adjacency matrix. We experimented with 0 to 3 convolutional layers, with shared layer parameters (Table \ref{tab:ablation_struct} row 2 to 5) and without weight sharing (Table \ref{tab:ablation_struct} rows 6 to 7). 0 convolution layer means directly classifying on the feature matrix $X$ in Eq. (1) rather than the refined feature matrix $Z$. Note, 0 convolution layer is different from concatenation. 0 layer still takes up a graphical structure with each object encoded separately, capturing the spatial relation with the pedestrian. The improvement from concatenation to 0 layer thus reflects the importance of spatial modeling.

The results suggest that a better blended graph (\ie,  more graph convolution layers) generally gives better performance, as evidenced by the gap from 0 to 1 to 2 layers. However, the performance gain seems to saturate, since the performances for 2 and 3 layers are similar. We hence choose 2 layers for lighter computation. Note also that here the 2-layer and 3-layer models have the same number of learnable parameters since the layer weights are shared. One may consider increasing model capacity by removing the weight sharing, however the experiments suggest that this seems to lead to overfitting the data and hence degrade the performance. The problem of overfitting is also reflected in row 8, where we increase the model capacity by relaxing the structure a fully-connected one. 

\textit{Choice of temporal connection}: since we consider intention prediction as a sequence modeling problem, it is not surprising that temporal connection factors heavily in the model (row 9). In addition, removing the temporal connection among pedestrian nodes hurts the performance (row 10), which means it is important for the pedestrian node to maintain a temporal history. However, adding more temporal connection may not be always beneficial, as evidenced by row 11 where an additional GRU was introduced on aggregated context features. The reason might be that the pedestrian GRU alone would suffice when the graph nodes are communicated sufficiently and the addition of context GRU may introduce redundancy.

\noindent\textbf{Choice of Features:} \label{sec:feat}
Given a chosen model structure, we experiment with incorporating different information in the hope of finding a rich yet lightweight feature representation and report the results in Table \ref{tab:ablation_feat}. The gap over the pedestrian-only variation (row 1) confirms the effectiveness of including context information. However, additionally adding into pedestrian pose did not help with the performance. This may be due to the quality of the poses, similar to the situation with the pose-only baseline (Table \ref{tab:main} row 1). Though theoretically the model can learn to ignore poorly predicted poses, in practice poses behave similarly to a source noise, making the learning task more challenging. It is also worth pointing out that the semantic labels of the objects may not be essential to the reasoning, as evidenced by row 4 of Table \ref{tab:ablation_feat}. We hypothesize that this is because the change of relative position already contains information that indicates the object type.

\begin{table}[]
    \caption{\textbf{Ablation study on graph design}: we compare different design choices by varying: 1) number of graph convolution layers; 2) whether to use weight sharing across layers or not ("\textit{no sharing}"); 3) whether to use a fully connected graph; 4) ways introduce temporal relation ("\textit{no temporal / pedestrian GRU}"). We also compare with a structure that simply concatenates the features for the pedestrian and the context ("\textit{Concat}"), which is often a strong baseline. The prediction covers 30 frames / 1 second into the future.}
    \label{tab:ablation_struct}
    \centering
    \begin{tabular}{c|p{3cm}|c|c}
        \hline
        \# & \textbf{Model} & \textbf{1-30 frames} & \textbf{30$^\text{th}$ frame} \\
        \hline
        1 & Concat & 76.96\% & 75.7\% \\ 
        \hline
        2 & 0 layer & 77.97\% & 76.38\% \\ 
        3 & 1 layer & 78.26\% & 76.68\% \\ 
        4 & 2 layer (\textbf{Ours}) & \textbf{79.28\%} & 76.98\% \\ 
        5 & 3 layers & 78.75\% & \textbf{77.28\%} \\ 
        \hline
        6 & 2 layers, no sharing & 78.17\% & 76.83\% \\ 
        7 & 3 layers, no sharing & 78.64\% & 76.83\% \\ 
        \hline
        8 & 2 layers, FC & 78.42\% & 76.53\% \\ 
        \hline
        9 & 2 layer, no temporal & 71.01\% & 69.96\% \\ 
        10 & 2 layer, no ped GRU & 78.58\% & 76.68\% \\ 
        11 & 2 layer, add ctxt GRU & 78.85\% & 78.62\% \\
        \hline
    \end{tabular}
\end{table}


\begin{table}[t]
    \caption{\textbf{Ablation study on the choice of features}: given fixed optimal structure, we test different feature choices. Adding context information shows to be useful. The prediction covers 30 frames, 1 second in the future (Section \ref{sec:feat}).}
    \label{tab:ablation_feat}
    \centering
    \begin{tabular}{c|l|c|c}
        \hline
        \textbf{\#} & \textbf{Model} & \textbf{1-30 frames} & \textbf{30$^\text{th}$ frame} \\
        \hline
        1 & Graph - Pedestrian & 77.81\% & 76.32\% \\ 
        2 & G - Ped + ctxt (\textbf{Ours}) & \textbf{79.28}\% & \textbf{76.98}\% \\ 
        3 & G - Ped + ctxt + pose & 76.11\% & 74.14\% \\ 
        4 & G - Ped + ctxt + objCls & 78,21\% & 76.14\% \\
        \hline
    \end{tabular}
\end{table}


\subsubsection{\rev{Extending the Temporal Horizon}}
Here, we would study how the prediction performance changes as we extend the temporal horizon. In addition to previous results on 30 frames, we extend prediction to 60 and 90 frames in this subsection, which corresponds to 2 and 3 seconds into the future. 

\begin{figure}
    \centering
    \includegraphics[width=\linewidth]{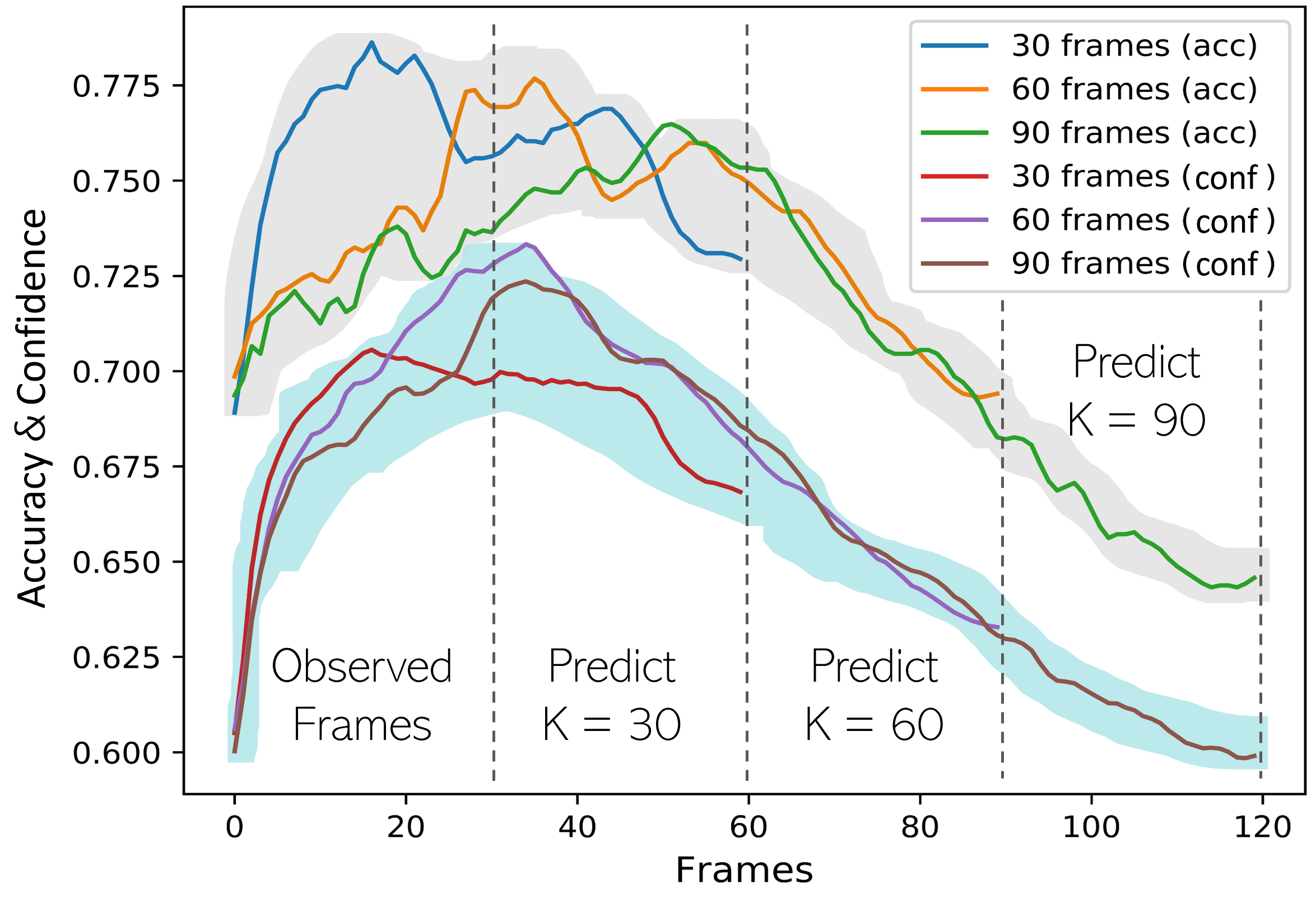} \vspace{-5pt}
    \caption{Prediction accuracy (gray shade) and confidence (light teal shade) as time horizon increases. We analyze three models which are trained to predict for up to 30, 60, and 90 frames into the future, each taking 1 second (frame-rate = 30 FPS) of observation as input. Specifically, ``\textit{x frames (acc/conf)}" refers to the accuracy or confidence curve for the model trained to predict over $x$ future frames.}
    \label{fig:horizon}
\end{figure}

Table \ref{tab:horizon} and Fig. \ref{fig:horizon} summarize the results. In general, prediction improves as more observations come in, as evidenced by the increasing curves in the left-most region in Fig. \ref{fig:horizon}. The turning point occurs when switching into prediction, where both the accuracy and confidence \rev{(the uncalibrated probability calculated as the sigmoid of the logits, the direct output of the last linear layer in the classifier)} decrease as the temporal horizon grows. The best accuracy at each region is achieved by models trained with that specific setting. However, it is interesting that the confidence may not be consistent with the performance; for example, the 30-frame model gets the best performance on the first 30 future frames while being overall least confident about these predictions. 
We gain more certainty when there are more frames of data available. As the 60 and 90 frame models receive more data and supervision they can infer about the intent more confidently. 

\begin{table}[]
    \caption{Accuracy compared at different prediction lengths. The predictions are over 30, 60, and 90 frames, or equivalently, 1, 2, or 3 seconds in the future.}
    \label{tab:horizon} 
    \centering
    \begin{tabular}{c|c|c}
        \hline
        \textbf{Length} & \textbf{Avg on 1-$K$ frames} & \textbf{On $K^\text{th}$ frame} \\
        \hline
        30 frames (1s) & 79.28\% & 76.98\% \\
        60 frames (2s) & 75.10\% & 73.09\% \\
        90 frames (3s) & 71.72\% & 68.31\% \\
        \hline
    \end{tabular}
\end{table}

\subsubsection{\rev{Results of Location-Centric Prediction}}



We restructure the dataset and obtain 32 video clips with 72882 frames, of which 14808 frames (around 20.3\%) contains the crossing behavior. The number of pedestrians per frame ranges from 0 to 9, with an average of 1.76 pedestrians. We use the same framework proposed in Section \ref{sec:method_gcn} as a lightweight learning scheme for prediction in the location-centric scenario. 

Similar to the pedestrian-centric setting, we compare with a concatenation baseline to demonstrate the effectiveness of the graph structure, with results shown in Table \ref{tab:loc_centric}. However, in this case we only train the concatenation model for one epoch, as opposed to experiments in Table \ref{tab:ablation_struct} where the model was trained to converge. This one epoch can be considered as a pretraining stage, and the location-centric graph then operates on the features extracted from the pretrained concatenation model. Note that the graph model is lighter to train with features as input, consuming about one-tenth of GPU memory and taking about one-fifth of time to complete an epoch.


\begin{table}[t]
    \caption{Results for the location-centric setting. With a pretrained concatenation model, the location-centric graph is able to continue the task learning in a memory and computation efficient manner. The prediction covers 30 frames, \ie, 1 second in the future.
    }
    \label{tab:loc_centric} 
    \centering
    \begin{tabular}{c|l|c|c}
        \hline
        \textbf{\#} & \textbf{Model} & \textbf{Avg on 1-30 frames} & \textbf{On 30$^\text{th}$ frame} \\ \hline
        1 & Concat & 74.13\% & 71.74\% \\ 
        2 & Graph & 86.38\% & 81.88\% \\
        \hline
    \end{tabular}
\end{table}

\subsection{Results on STIP Dataset}
Similar to the settings on the JAAD dataset, we  examine how the performance of our model scales as the time horizon increases. Possibly due to denser pedestrian appearances, the baseline \textit{concat} model encountered memory issue with longer time horizon. We therefore train the \textit{concat} model under a setting with short observing and predicting time, and use its weights to initialize the graph model as before. The graph model is then fine-tuned to predict for longer temporal horizon into the future. \rev{The results are reported in Table \ref{tab:stip_horizon_2}. With the data from only the front camera,} predicting for 3 seconds in the future achieved slightly better accuracy than predicting for 2 seconds. However, the confidence of predictions at each step decreases monotonically overtime. \rev{A different trend is observed when all three cameras are used. In this case, objects from all three cameras are considered in a single graph, with the locations for side-camera objects adjusted by shifting them to the left or right (shifting the $x$ coordinates) based on the view that the objects are from. When using all three cameras, having observed 2 seconds predicts no worse than having observed 4 seconds. This is in accordance with the assumption that the side cameras give a wider view of the scene and provide better cues for relationship reasoning to predict pedestrian intent, which is also verified by the boost in performance across most of the settings.}

\begin{table}[]
    \caption{\rev{Accuracy compared at different prediction lengths on STIP dataset. Our model takes in 2 or 4 seconds of observation, and predicts for 1, 2, or 3 seconds into the future. Second column only reports results of spatiotemporal reasoning on a graph built from the front camera, and the last column shows results of all three cameras (left, front, \& right).}}
    \label{tab:stip_horizon_2} 
    \centering
    \rev{
    \begin{tabular}{c|c|c}
        \hline
        \textbf{Length} & \multicolumn{2}{c}{\textbf{Avg Accuracy on predicted frames}} \\ 
        \cline{2-3} ~ & \textbf{Front Camera} & \textbf{All 3 Cameras} \\
        \hline
        2s $\rightarrow$ 1s & 78.68\% & 81.20\% \\ 
        2s $\rightarrow$ 2s & 78.09\% & 80.49\% \\ 
        2s $\rightarrow$ 3s & 78.16\% & 80.77\% \\ 
        4s $\rightarrow$ 1s & 80.36\% & 81.53\% \\ 
        4s $\rightarrow$ 2s & 80.06\% & 81.73\% \\ 
        4s $\rightarrow$ 3s & 80.32\% & 79.62\% \\ 
        \hline
    \end{tabular}
    }
\end{table}



\section{Conclusion}
In this paper, we proposed a method based on graph convolution to model the spatiotemporal relationships of the pedestrians and other objects in the scene. We build our spatiotemporal graph by considering each segmented instance in a frame as a node. Pedestrian-centric and location-centric graphs are built and extracted features for each graph at each time point is fed into a gated recurrent unit. We use this model for predicting the pedestrians intention, which is defined as the future actions of cross or not-cross. We also introduced a dataset, STIP, tailored for intent prediction in dense driving scenes. The results show that our spatiotemporal relationship reasoning model can predict the intention with an accuracy of over 80\% on STIP and a bit less than 80\% on JAAD about one second ahead of the time that the actual crossing happens. As a direction for future work, further improvements over the spatiotemporal reasoning framework can be obtained by incorporating more intermediate annotations \rev{as well as a probability calibration method \cite{guo2017calibration} for obtaining reliable confidence scores for the predictions}.  

\section*{Acknowledgments}
This research was supported by the Toyota Research Institute (TRI). This article solely reflects the opinions and conclusions of its authors and not TRI or any other Toyota entity. The authors would like to thank Karttikeya Mangalam for helping with obtaining results of the baseline methods.


{
\bibliographystyle{ieeetr}
\bibliography{egbib}

\begin{thebibliography}{10}

\bibitem{cite:Springer14VB}
D.~Ger{\'o}nimo and A.~M. L{\'o}pez, {\em Vision-based pedestrian protection
  systems for intelligent vehicles}.
\newblock Springer, 2014.

\bibitem{cite:TPAMI18Towards}
S.~Zhang, R.~Benenson, M.~Omran, J.~Hosang, and B.~Schiele, ``Towards reaching
  human performance in pedestrian detection,'' {\em TPAMI}, vol.~40, no.~4,
  pp.~973--986, 2018.

\bibitem{cite:ICIP17Simple}
N.~Wojke, A.~Bewley, and D.~Paulus, ``Simple online and realtime tracking with
  a deep association metric,'' in {\em ICIP}, pp.~3645--3649, 2017.

\bibitem{cite:TPAMI14VirtualReal}
D.~Vazquez, A.~M. Lopez, J.~Marin, D.~Ponsa, and D.~Geronimo, ``Virtual and
  real world adaptation for pedestrian detection,'' {\em TPAMI}, vol.~36,
  no.~4, pp.~797--809, 2014.

\bibitem{cite:CVPR18OcclusionSingle}
J.~Noh, S.~Lee, B.~Kim, and G.~Kim, ``Improving occlusion and hard negative
  handling for single-stage pedestrian detectors,'' in {\em CVPR}, June 2018.

\bibitem{cite:ECCV18SingleAsymLF}
W.~Liu, S.~Liao, W.~Hu, X.~Liang, and X.~Chen, ``Learning efficient
  single-stage pedestrian detectors by asymptotic localization fitting,'' in
  {\em ECCV}, September 2018.

\bibitem{cite:ECCV18Crowd}
S.~Zhang, L.~Wen, X.~Bian, Z.~Lei, and S.~Z. Li, ``Occlusion-aware {R-CNN}:
  Detecting pedestrians in a crowd,'' in {\em ECCV}, September 2018.

\bibitem{cite:CVPR18Crowd}
X.~Wang, T.~Xiao, Y.~Jiang, S.~Shao, J.~Sun, and C.~Shen, ``Repulsion loss:
  Detecting pedestrians in a crowd,'' in {\em CVPR}, June 2018.

\bibitem{cite:ITSC14BLAC}
R.~Quintero, I.~Parra, D.~F. Llorca, and M.~Sotelo, ``Pedestrian path
  prediction based on body language and action classification,'' in {\em ITSC},
  pp.~679--684, 2014.

\bibitem{cite:ECCV14CBP}
J.~F.~P. Kooij, N.~Schneider, F.~Flohr, and D.~M. Gavrila, ``Context-based
  pedestrian path prediction,'' in {\em ECCV}, pp.~618--633, 2014.

\bibitem{cite:NIPS17CASNSC}
N.~Japuria, G.~Habibi, and J.~P. How, ``{CASNSC}: A context-based approach for
  accurate pedestrian motion prediction at intersections,'' in {\em NeurIPS},
  2017.

\bibitem{cite:ArxivLTRPM}
K.~Saleh, M.~Hossny, and S.~Nahavandi, ``Long-term recurrent predictive model
  for intent prediction of pedestrians via inverse reinforcement learning,'' in
  {\em DICTA}, pp.~1--8, 2018.

\bibitem{cite:ICRA16IALT}
V.~Karasev, A.~Ayvaci, B.~Heisele, and S.~Soatto, ``Intent-aware long-term
  prediction of pedestrian motion,'' in {\em ICRA}, pp.~2543--2549, 2016.

\bibitem{cite:Arxiv18Transferable}
N.~Jaipuria, G.~Habibi, and J.~P. How, ``A transferable pedestrian motion
  prediction model for intersections with different geometries,'' {\em arXiv
  preprint arXiv:1806.09444}, 2018.

\bibitem{mangalam2020disentangling}
K.~Mangalam, E.~Adeli, K.-H. Lee, A.~Gaidon, and J.~C. Niebles, ``Disentangling
  human dynamics for pedestrian locomotion forecasting with noisy
  supervision,'' in {\em WACV}, 2020.

\bibitem{cite:CVPR18CrowdTraj}
Y.~Xu, Z.~Piao, and S.~Gao, ``Encoding crowd interaction with deep neural
  network for pedestrian trajectory prediction,'' in {\em CVPR}, June 2018.

\bibitem{alahi2016social}
A.~Alahi, K.~Goel, V.~Ramanathan, A.~Robicquet, L.~Fei-Fei, and S.~Savarese,
  ``{Social LSTM}: Human trajectory prediction in crowded spaces,'' in {\em
  CVPR}, pp.~961--971, 2016.

\bibitem{sadeghian2018sophie}
A.~Sadeghian, V.~Kosaraju, A.~Sadeghian, N.~Hirose, and S.~Savarese, ``Sophie:
  An attentive gan for predicting paths compliant to social and physical
  constraints,'' {\em arXiv preprint arXiv:1806.01482}, 2018.

\bibitem{cite:ER13IAPA}
T.~Bandyopadhyay, C.~Z. Jie, D.~Hsu, M.~H. Ang, D.~Rus, and E.~Frazzoli,
  ``Intention-aware pedestrian avoidance,'' in {\em Experimental Robotics},
  pp.~963--977, 2013.

\bibitem{cite:ITSC14MCBM}
S.~Bonnin, T.~H. Weisswange, F.~Kummert, and J.~Schm{\"u}dderich, ``Pedestrian
  crossing prediction using multiple context-based models,'' in {\em ITSC},
  pp.~378--385, 2014.

\bibitem{cite:IVS182DPose}
Z.~Fang and A.~M. L{\'o}pez, ``Is the pedestrian going to cross? answering by
  {2D} pose estimation,'' in {\em 2018 IEEE Intelligent Vehicles Symposium
  (IV)}, pp.~1271--1276, IEEE, 2018.

\bibitem{seo2018structured}
Y.~Seo, M.~Defferrard, P.~Vandergheynst, and X.~Bresson, ``Structured sequence
  modeling with graph convolutional recurrent networks,'' in {\em International
  Conference on Neural Information Processing}, pp.~362--373, 2018.

\bibitem{rasouli2019pie}
A.~Rasouli, I.~Kotseruba, T.~Kunic, and J.~K. Tsotsos, ``{PIE}: A large-scale
  dataset and models for pedestrian intention estimation and trajectory
  prediction,'' in {\em ICCV}, pp.~6262--6271, 2019.

\bibitem{li2018learning}
J.~Li, A.~Raventos, A.~Bhargava, T.~Tagawa, and A.~Gaidon, ``Learning to fuse
  things and stuff,'' {\em arXiv preprint arXiv:1812.01192}, 2018.

\bibitem{cite:ECCV18OCcclusion}
C.~Zhou and J.~Yuan, ``Bi-box regression for pedestrian detection and occlusion
  estimation,'' in {\em ECCV}, September 2018.

\bibitem{cite:CVPR18Occlusion}
S.~Zhang, J.~Yang, and B.~Schiele, ``Occluded pedestrian detection through
  guided attention in cnns,'' in {\em CVPR}, June 2018.

\bibitem{cite:ECCV16MultiTrack}
S.~Tang, B.~Andres, M.~Andriluka, and B.~Schiele, ``Multi-person tracking by
  multicut and deep matching,'' in {\em ECCV}, pp.~100--111, 2016.

\bibitem{cite:CVPR17MultiTrack}
E.~Insafutdinov, M.~Andriluka, L.~Pishchulin, S.~Tang, E.~Levinkov, B.~Andres,
  and B.~Schiele, ``{ArtTrack}: Articulated multi-person tracking in the
  wild,'' in {\em CVPR}, July 2017.

\bibitem{cite:CVPR17MultiTrackPose}
U.~Iqbal, A.~Milan, and J.~Gall, ``{PoseTrack}: Joint multi-person pose
  estimation and tracking,'' in {\em CVPR}, July 2017.

\bibitem{cite:ECCV18MultiTrackPose}
D.~Yu, K.~Su, J.~Sun, and C.~Wang, ``Multi-person pose estimation for pose
  tracking with enhanced cascaded pyramid network,'' in {\em ECCV},
  pp.~221--226, 2018.

\bibitem{cite:Arxiv18TrackPose}
Y.~Xiu, J.~Li, H.~Wang, Y.~Fang, and C.~Lu, ``Pose flow: Efficient online pose
  tracking,'' {\em CoRR}, vol.~abs/1802.00977, 2018.

\bibitem{cite:CVPR17MultiTrackReid}
S.~Tang, M.~Andriluka, B.~Andres, and B.~Schiele, ``Multiple people tracking by
  lifted multicut and person re-identification,'' in {\em CVPR}, July 2017.

\bibitem{cite:CVPR18MultiTrackReid}
E.~Ristani and C.~Tomasi, ``Features for multi-target multi-camera tracking and
  re-identification,'' in {\em CVPR}, June 2018.

\bibitem{cite:ICRA16ADL}
Y.~F. Chen, M.~Liu, and J.~P. How, ``Augmented dictionary learning for motion
  prediction,'' in {\em ICRA}, pp.~2527--2534, 2016.

\bibitem{gupta2018social}
A.~Gupta, J.~Johnson, L.~Fei-Fei, S.~Savarese, and A.~Alahi, ``{Social GAN}:
  Socially acceptable trajectories with generative adversarial networks,'' in
  {\em CVPR}, pp.~2255--2264, 2018.

\bibitem{lan2014hierarchical}
T.~Lan, T.-C. Chen, and S.~Savarese, ``A hierarchical representation for future
  action prediction,'' in {\em ECCV}, pp.~689--704, 2014.

\bibitem{xie2017learning}
D.~Xie, T.~Shu, S.~Todorovic, and S.-C. Zhu, ``Learning and inferring “dark
  matter” and predicting human intents and trajectories in videos,'' {\em
  TPAMI}, vol.~40, no.~7, pp.~1639--1652, 2017.

\bibitem{wei2018and}
P.~Wei, Y.~Liu, T.~Shu, N.~Zheng, and S.-C. Zhu, ``Where and why are they
  looking? jointly inferring human attention and intentions in complex tasks,''
  in {\em CVPR}, pp.~6801--6809, 2018.

\bibitem{abu2018will}
Y.~Abu~Farha, A.~Richard, and J.~Gall, ``When will you do what?-anticipating
  temporal occurrences of activities,'' in {\em CVPR}, pp.~5343--5352, 2018.

\bibitem{shi2018action}
Y.~Shi, B.~Fernando, and R.~Hartley, ``Action anticipation with rbf kernelized
  feature mapping {RNN},'' in {\em ECCV}, pp.~301--317, 2018.

\bibitem{rhinehart2017first}
N.~Rhinehart and K.~M. Kitani, ``First-person activity forecasting with online
  inverse reinforcement learning,'' in {\em ICCV}, pp.~3696--3705, 2017.

\bibitem{koppula2016anticipating}
H.~S. Koppula and A.~Saxena, ``Anticipating human activities using object
  affordances for reactive robotic response,'' {\em TPAMI}, vol.~38, no.~1,
  pp.~14--29, 2016.

\bibitem{chen2018part}
L.~Chen, J.~Lu, Z.~Song, and J.~Zhou, ``Part-activated deep reinforcement
  learning for action prediction,'' in {\em ECCV}, pp.~421--436, 2018.

\bibitem{battaglia2016interaction}
P.~Battaglia, R.~Pascanu, M.~Lai, D.~J. Rezende, {\em et~al.}, ``Interaction
  networks for learning about objects, relations and physics,'' in {\em
  NeurIPS}, pp.~4502--4510, 2016.

\bibitem{van2018relational}
S.~Van~Steenkiste, M.~Chang, K.~Greff, and J.~Schmidhuber, ``Relational neural
  expectation maximization: Unsupervised discovery of objects and their
  interactions,'' {\em arXiv preprint arXiv:1802.10353}, 2018.

\bibitem{xu2017scene}
D.~Xu, Y.~Zhu, C.~B. Choy, and L.~Fei-Fei, ``Scene graph generation by
  iterative message passing,'' in {\em CVPR}, pp.~5410--5419, 2017.

\bibitem{zellers2018neural}
R.~Zellers, M.~Yatskar, S.~Thomson, and Y.~Choi, ``Neural motifs: Scene graph
  parsing with global context,'' in {\em CVPR}, pp.~5831--5840, 2018.

\bibitem{yang2018graph}
J.~Yang, J.~Lu, S.~Lee, D.~Batra, and D.~Parikh, ``Graph r-cnn for scene graph
  generation,'' in {\em ECCV}, pp.~670--685, 2018.

\bibitem{cong2018scene}
W.~Cong, W.~Wang, and W.-C. Lee, ``Scene graph generation via conditional
  random fields,'' {\em arXiv preprint arXiv:1811.08075}, 2018.

\bibitem{ibrahim2016hierarchical}
M.~S. Ibrahim, S.~Muralidharan, Z.~Deng, A.~Vahdat, and G.~Mori, ``A
  hierarchical deep temporal model for group activity recognition,'' in {\em
  CVPR}, pp.~1971--1980, 2016.

\bibitem{johnson2018image}
J.~Johnson, A.~Gupta, and L.~Fei-Fei, ``Image generation from scene graphs,''
  in {\em CVPR}, pp.~1219--1228, 2018.

\bibitem{wang2018videos}
X.~Wang and A.~Gupta, ``Videos as space-time region graphs,'' in {\em ECCV},
  pp.~399--417, 2018.

\bibitem{kipf2018neural}
T.~Kipf, E.~Fetaya, K.-C. Wang, M.~Welling, and R.~Zemel, ``Neural relational
  inference for interacting systems,'' {\em arXiv preprint arXiv:1802.04687},
  2018.

\bibitem{teney2017graph}
D.~Teney, L.~Liu, and A.~van~den Hengel, ``Graph-structured representations for
  visual question answering,'' in {\em CVPR}, pp.~1--9, 2017.

\bibitem{shi2018explainable}
J.~Shi, H.~Zhang, and J.~Li, ``Explainable and explicit visual reasoning over
  scene graphs,'' in {\em AAAI}, 2018.

\bibitem{aditya2018image}
S.~Aditya, Y.~Yang, C.~Baral, Y.~Aloimonos, and C.~Ferm{\"u}ller, ``Image
  understanding using vision and reasoning through scene description graph,''
  {\em CVIU}, vol.~173, pp.~33--45, 2018.

\bibitem{chen2019graph}
Y.~Chen, M.~Rohrbach, Z.~Yan, S.~Yan, J.~Feng, and Y.~Kalantidis, ``Graph-based
  global reasoning networks,'' in {\em CVPR}, 2019.

\bibitem{jain2016structural}
A.~Jain, A.~R. Zamir, S.~Savarese, and A.~Saxena, ``{Structural-RNN}: Deep
  learning on spatio-temporal graphs,'' in {\em CVPR}, pp.~5308--5317, 2016.

\bibitem{qi2018learning}
S.~Qi, W.~Wang, B.~Jia, J.~Shen, and S.-C. Zhu, ``Learning human-object
  interactions by graph parsing neural networks,'' in {\em ECCV}, pp.~401--417,
  2018.

\bibitem{shu2017cern}
T.~Shu, S.~Todorovic, and S.-C. Zhu, ``Cern: confidence-energy recurrent
  network for group activity recognition,'' in {\em CVPR}, pp.~5523--5531,
  2017.

\bibitem{ren2015faster}
S.~Ren, K.~He, R.~Girshick, and J.~Sun, ``Faster {R-CNN}: Towards real-time
  object detection with region proposal networks,'' in {\em NeurIPS},
  pp.~91--99, 2015.

\bibitem{lin2017focal}
T.-Y. Lin, P.~Goyal, R.~Girshick, K.~He, and P.~Doll{\'a}r, ``Focal loss for
  dense object detection,'' in {\em CVPR}, pp.~2980--2988, 2017.

\bibitem{redmon2018yolov3}
J.~Redmon and A.~Farhadi, ``Yolov3: An incremental improvement,'' {\em arXiv
  preprint arXiv:1804.02767}, 2018.

\bibitem{neuhold2017mapillary}
G.~Neuhold, T.~Ollmann, S.~Rota~Bulo, and P.~Kontschieder, ``The mapillary
  vistas dataset for semantic understanding of street scenes,'' in {\em ICCV},
  pp.~4990--4999, 2017.

\bibitem{kipf2016semi}
T.~N. Kipf and M.~Welling, ``Semi-supervised classification with graph
  convolutional networks,'' {\em arXiv preprint arXiv:1609.02907}, 2016.

\bibitem{zeng2017agent}
K.-H. Zeng, S.-H. Chou, F.-H. Chan, J.~Carlos~Niebles, and M.~Sun,
  ``Agent-centric risk assessment: Accident anticipation and risky region
  localization,'' in {\em CVPR}, pp.~2222--2230, 2017.

\bibitem{kotseruba2016joint}
I.~Kotseruba, A.~Rasouli, and J.~K. Tsotsos, ``Joint attention in autonomous
  driving (jaad),'' {\em arXiv preprint arXiv:1609.04741}, 2016.

\bibitem{geiger2013vision}
A.~Geiger, P.~Lenz, C.~Stiller, and R.~Urtasun, ``Vision meets robotics: The
  kitti dataset,'' {\em IJRR}, vol.~32, no.~11, pp.~1231--1237, 2013.

\bibitem{yu2018bdd100k}
F.~Yu, W.~Xian, Y.~Chen, F.~Liu, M.~Liao, V.~Madhavan, and T.~Darrell,
  ``{BDD100k}: A diverse driving video database with scalable annotation
  tooling,'' {\em arXiv preprint arXiv:1805.04687}, 2018.

\bibitem{kim2019pedx}
W.~Kim, M.~S. Ramanagopal, C.~Barto, M.-Y. Yu, K.~Rosaen, N.~Goumas,
  R.~Vasudevan, and M.~Johnson-Roberson, ``{PedX}: Benchmark dataset for metric
  3-d pose estimation of pedestrians in complex urban intersections,'' {\em
  IEEE RA-L}, vol.~4, no.~2, pp.~1940--1947, 2019.

\bibitem{caesar2019nuscenes}
H.~Caesar, V.~Bankiti, A.~H. Lang, S.~Vora, V.~E. Liong, Q.~Xu, A.~Krishnan,
  Y.~Pan, G.~Baldan, and O.~Beijbom, ``{nuScenes}: A multimodal dataset for
  autonomous driving,'' {\em arXiv preprint arXiv:1903.11027}, 2019.

\bibitem{hamid2015joint}
S.~Hamid~Rezatofighi, A.~Milan, Z.~Zhang, Q.~Shi, A.~Dick, and I.~Reid, ``Joint
  probabilistic data association revisited,'' in {\em ICCV}, pp.~3047--3055,
  2015.

\bibitem{cao2018openpose}
Z.~Cao, G.~Hidalgo, T.~Simon, S.-E. Wei, and Y.~Sheikh, ``Open{P}ose: realtime
  multi-person 2{D} pose estimation using {P}art {A}ffinity {F}ields,'' in {\em
  arXiv preprint arXiv:1812.08008}, 2018.

\bibitem{lin2014microsoft}
T.-Y. Lin, M.~Maire, S.~Belongie, J.~Hays, P.~Perona, D.~Ramanan,
  P.~Doll{\'a}r, and C.~L. Zitnick, ``Microsoft coco: Common objects in
  context,'' in {\em ECCV}, pp.~740--755, 2014.

\bibitem{TSN2016ECCV}
L.~Wang, Y.~Xiong, Z.~Wang, Y.~Qiao, D.~Lin, X.~Tang, and L.~{Val Gool},
  ``Temporal segment networks: Towards good practices for deep action
  recognition,'' in {\em ECCV}, 2016.

\bibitem{TRN}
B.~Zhou, A.~Andonian, A.~Oliva, and A.~Torralba, ``Temporal relational
  reasoning in videos,'' in {\em ECCV}, pp.~803--818, 2018.

\bibitem{guo2017calibration}
C.~Guo, G.~Pleiss, Y.~Sun, and K.~Q. Weinberger, ``On calibration of modern
  neural networks,'' in {\em ICML}, pp.~1321--1330, 2017.

\end{thebibliography}
}

\end{document}